# Pyramid hybrid pooling quantization for efficient fine-grained image retrieval

Ziyun Zeng [a,d,1], Jinpeng Wang [a,d,1], Bin Chen [b,d,*], Tao Dai [c], Shu-Tao Xia [a,d], Zhi Wang [a]

[a] Tsinghua Shenzhen International Graduate School, Tsinghua University, China
[b] Harbin Institute of Technology, Shenzhen, China
[c] Shenzhen University, China
[d] Research Center of Artificial Intelligence, Peng Cheng Laboratory, China

ABSTRACT

Deep hashing approaches, including deep quantization and deep binary hashing, have become a common solution to large-scale image retrieval due to their high computation and storage efficiency. Most existing hashing methods cannot produce satisfactory results for fine-grained retrieval, because they usually adopt the outputs of the last CNN layer to generate binary codes. Since deeper layers tend to summarize visual clues, *e.g.*, texture, into abstract semantics, *e.g.*, dogs and cats, the feature produced by the last CNN layer is less effective in capturing subtle but discriminative visual details that mostly exist in shallow layers. To improve fine-grained image hashing, we propose ***P****yramid* ***H****ybrid* ***P****ooling* ***Q****uantization* (**PHPQ**). Specifically, we propose a ***P****yramid* ***H****ybrid* ***P****ooling* (**PHP**) module to capture and preserve fine-grained semantic information from multi-level features, which emphasizes the subtle discrimination of different sub-categories. Besides, we propose a learnable quantization module with a *partial codebook attention mechanism*, which helps to optimize the most relevant codewords and improves the quantization. Comprehensive experiments on two widely-used public benchmarks, *i.e.*, CUB-200-2011 and Stanford Dogs, demonstrate that PHPQ outperforms state-of-the-art methods.

## 1. Introduction

The explosively growing amount of images on the web raises the concern of search efficiency. Due to high efficiency in both computation and storage, hashing [1,2] has become a popular solution for large-scale retrieval. In general, the hashing methods can be categorized into two series, *i.e.*, binary hashing [2,3], and quantization [4,5]. Binary hashing methods project high-dimensional data into short binary codes so that pairwise similarity can be quickly computed by Hamming distance. Quantization methods divide the original feature space into disjoint subspaces and approximately represent the data points in each subspace by its centroid. By pre-computing a look-up table of inter-centroid similarities, the similarity between any data pair can be quickly obtained.

Deep learning has become a common recipe for hashing methods in the past few years. Note that most existing deep hashing methods [6,7] are designed for *coarse-grained* retrieval. They can only support image retrieval from general concepts, *e.g.*, dogs or birds, which may fall short of practical needs. As a result, hashing for *fine-grained* retrieval becomes an important demand. We illustrate fine-grained samples of typical datasets [8,9] in Fig. 1, where one can observe that samples belonging to distinct categories exhibit nuanced inter-class distinctions, whereas samples within the same category display significant intra-class variations. Consequently, fine-grained image hashing requires discrimination among sub-categories in a meta-category, such as different varieties of dogs. Therefore, it is more challenging to learn representations that keep robust to intra-class variances (*e.g.*, various viewpoints and backgrounds) and discriminative to subtle inter-class differences. Unfortunately, coarse-grained hashing methods merely utilize the outputs of the last CNN feature layer to generate binary codes. They may lose visual details in images and produce inferior representations for fine-grained retrieval. On the other hand, recent fine-grained image hashing methods add heavy modules for region localization [10,11] or local feature alignment [12]. It is not surprising that these modules can help to capture fine-grained features while they increase both memory and computation overhead. Besides, some of these methods require additional annotations, such as bounding boxes, making it labor-intensive for application.

* Corresponding author.
*E-mail addresses:* zengzy21@mails.tsinghua.edu.cn (Z. Zeng), wjp20@mails.tsinghua.edu.cn (J. Wang), chenbin2021@hit.edu.cn (B. Chen), daitao@szu.edu.cn (T. Dai), xiast@sz.tsinghua.edu.cn (S.-T. Xia), wangzhi@sz.tsinghua.edu.cn (Z. Wang).

[1] Z. Zeng and J. Wang contributed equally to this work.

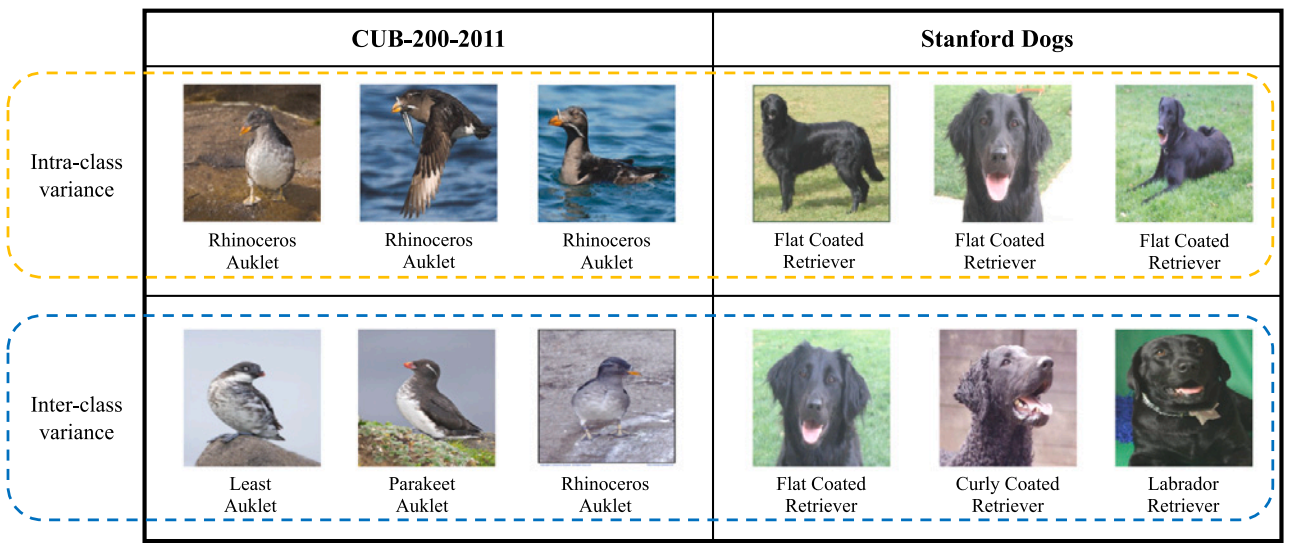

**Fig. 1.** Example fine-grained categories in CUB-200-2011 [8] and Stanford Dogs [9]. Samples within the same category display significant intra-class variances, whereas samples belonging to distinct categories exhibit nuanced inter-class variances.

To improve the hashing for fine-grained image retrieval, we propose ***P**yramid **H**ybrid **P**ooling **Q**uantization* (**PHPQ**). PHPQ is a quantization method that keeps more flexibility than binary hashing methods. While both quantization and binary hashing involve the storage of bit-wise codes, it is important to note that quantization computes similarities using real-space codewords. This utilization of real-space codewords contributes to a larger cardinality, thereby enhancing representation capability compared to the binary Hamming space [13]. In PHPQ, we design a lightweight *Pyramid Hybrid Pooling (PHP) module* to extract multi-level semantic information without additional annotations. The PHP module first utilizes pyramid features [14] from multiple CNN layers, each containing visual semantic information at a specific scale. Then, it fuses them through a tunable hybrid pooling strategy, which helps to focus and preserve critical semantic information from different visual scales. Besides, we propose a novel *partial codebook attention mechanism* to enhance quantization learning. Recently, [7] applied the attention mechanism to relax *hard* (*i.e.*, nearest neighbor-based) codeword selection. The *soft* quantized representation for an image is the weighted sum of all codewords, where the attention weights are bounded and kept non-zero. Although the encoding process becomes differentiable, the optimization is less effective as *all* codewords, including those irrelevant ones, are updated in backpropagation. Since irrelevant codewords always receive perturbation, it degrades quantized representations and harms fine-grained retrieval. By contrast, our partial attention mechanism filters out irrelevant codewords in the selection, guiding the optimization to those relevant ones attentively. Extensive experiments demonstrate that the partial attention mechanism improves deep quantization for fine-grained retrieval.

To summarize, we make the following contributions.

- We propose a deep quantization model for fine-grained image retrieval, which surpasses state-of-the-art methods.
- We propose a ***P**yramid **H**ybrid **P**ooling* (PHP) module to capture and preserve fine-grained semantic information, which raises the discriminability of representations.
- We propose a *partial codebook attention mechanism* to improve learnable quantization. It can guide the optimization to relevant codewords attentively, leading to better quantized representations for fine-grained image retrieval.

## 2. Related work

**Deep Quantization**: Traditional quantization methods [4,5] directly adopt handcrafted features and achieve suboptimal performance. Recently, several deep quantization methods [7,13,15–17] with CNNs [18,19] are proposed and achieve state-of-the-art performance. These works explore relational information among the training samples to learn semantic-preserved quantization. For instance, DTQ [15] aims to bring quantized positive samples closer while pushing quantized negative samples away using a triplet loss. PQN [7] further improves DTQ by incorporating an asymmetrical triplet loss, where the anchor representation remains unquantized to enhance discrimination. WSDAQ [16] and MeCoQ [13] achieve the same goal through contrastive learning [20]. CSQ [17] learns semantic hash centers and determines the quantized representation by referring to the closest center through a voting mechanism. Nevertheless, they are designed for coarse-grained image retrieval tasks, leaving fine-grained quantization under-explored. Our PHPQ attempts to design a quantization scheme under the fine-grained scenario. Besides, we further investigate the issues in learnable quantization [7,13] and propose a novel partial codebook attention mechanism to improve end-to-end quantization learning.

**Fine-Grained Image Hashing**: In recent years, several deep hashing methods have been proposed to improve fine-grained image retrieval. For example, SRLH [10] learns region localization and hash coding in a mutually reinforced way. CFH [11] utilizes the cross-modal correlation between semantic information and visual features to localize discriminative regions. ExchNet [12] aligns local features through a feature exchanging operation, producing discriminative part-level features while keeping the consistency of semantic information. Note that these methods rely on heavy sub-networks and thus increase the memory and computation overheads. Besides, some of them (*e.g.*, ExchNet) require additional annotations for fine-grained visual localization, making them expensive for real-world applications. On the other hand, FPH [21] aggregates different pyramid feature maps from the convolution layers by average pooling, which keeps the efficiency. However, the average pooling is weak in extracting fine-grained features because it equally treats informative and noisy areas on the feature maps.

In contrast, PHPQ does not require additional annotations or heavy sub-networks. It extracts fine-grained features through a lightweight pyramid hybrid pooling module, where the pooling schemes are level-specific. Therefore, it can efficiently capture discriminative semantic information and generate better binary codes for fine-grained retrieval.

## 3. Method

### 3.1. Problem definition

Given a labeled training set $\{(\boldsymbol{I}_n, y_n)\}_{n=1}^{N}$, where $\boldsymbol{I}_n \in \mathbb{R}^P$ denotes the $P$-dimensional flattened image vector, $N$ denotes the training set size, $y_n \in \{1, \ldots, N_c\}$ denotes the fine-grained category label for this image, and $N_c$ is the category number. The goal of quantization is to learn a quantizer $q : \mathbb{R}^P \mapsto \{-1, 1\}^{L_B}$ that maps the raw image vector $\boldsymbol{I}_n$ to a $L_B$-dimensional binary code $\boldsymbol{b}_n \in \{-1, 1\}^{L_B}$ and that preserves semantic information. To this end, we propose ***P**yramid **H**ybrid **P**ooling **Q**uantization* (**PHPQ**). Fig. 2 illustrates the framework of PHPQ. Initially, we extract hierarchical image features using a conventional CNN. These features undergo aggregation through our newly introduced PHP module. Following a projection layer, the image embedding is quantized using a learned quantization module, which is further enhanced by our novel partial codebook attention mechanism. The details are given in the next sections.

### 3.2. Pyramid hybrid pooling module

**Generalized Spatial Pooling**: Spatial pooling [22] is an efficient way to produce powerful descriptors from the convolutional feature maps. There are two representative methods in spatial pooling, *i.e.*, the Global Maximum Pooling (GMP) [23] and the Global Average Pooling (GAP) [24]. However, neither of them is powerful enough to capture discriminative fine-grained details, because: (**i**) GMP focuses on the highest response elements in the feature map while ignoring other small but discriminative regions to fine-grained recognition. (**ii**) GAP equally treats all elements, which produces less discriminative descriptors with noisy information from the background.

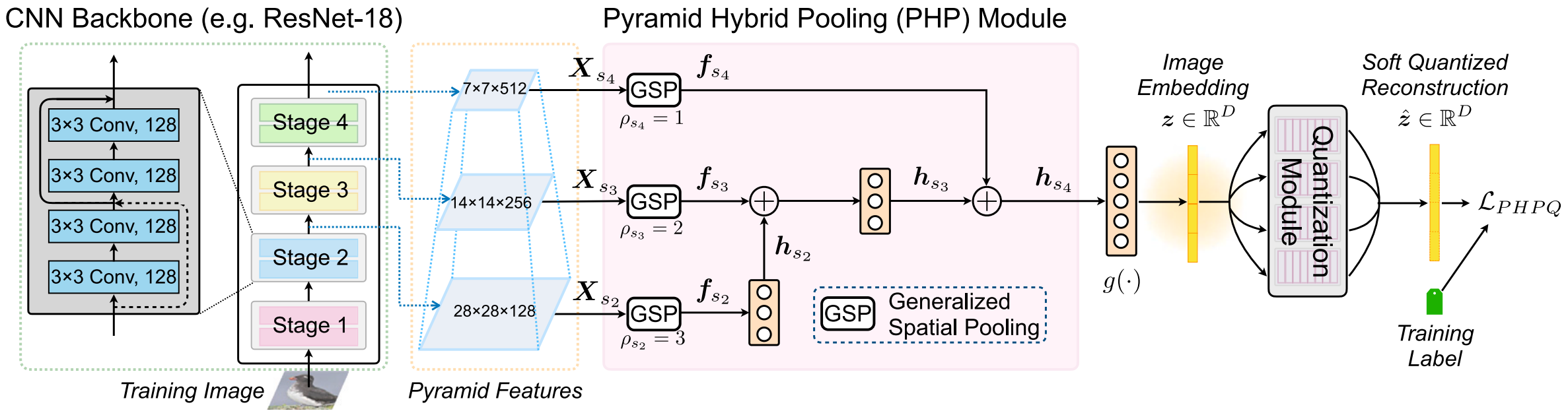

**Fig. 2.** The training framework of the ***Pyramid Hybrid Pooling Quantization*** (**PHPQ**). First, we send the raw image into a standard CNN, *e.g.*, ResNet-18, and extract pyramid features from different stages. Then, we aggregate these multi-level features by the proposed ***Pyramid Hybrid Pooling (PHP)*** *module and get the image embedding through a transformation layer*. Next, the image embedding will be quantized and reconstructed through the quantization module, where we propose a *Partial Codebook Attention Mechanism* to enhance end-to-end quantization learning. Finally, we define a learning objective with the soft quantized reconstructed embedding and optimize the model by back-propagation.

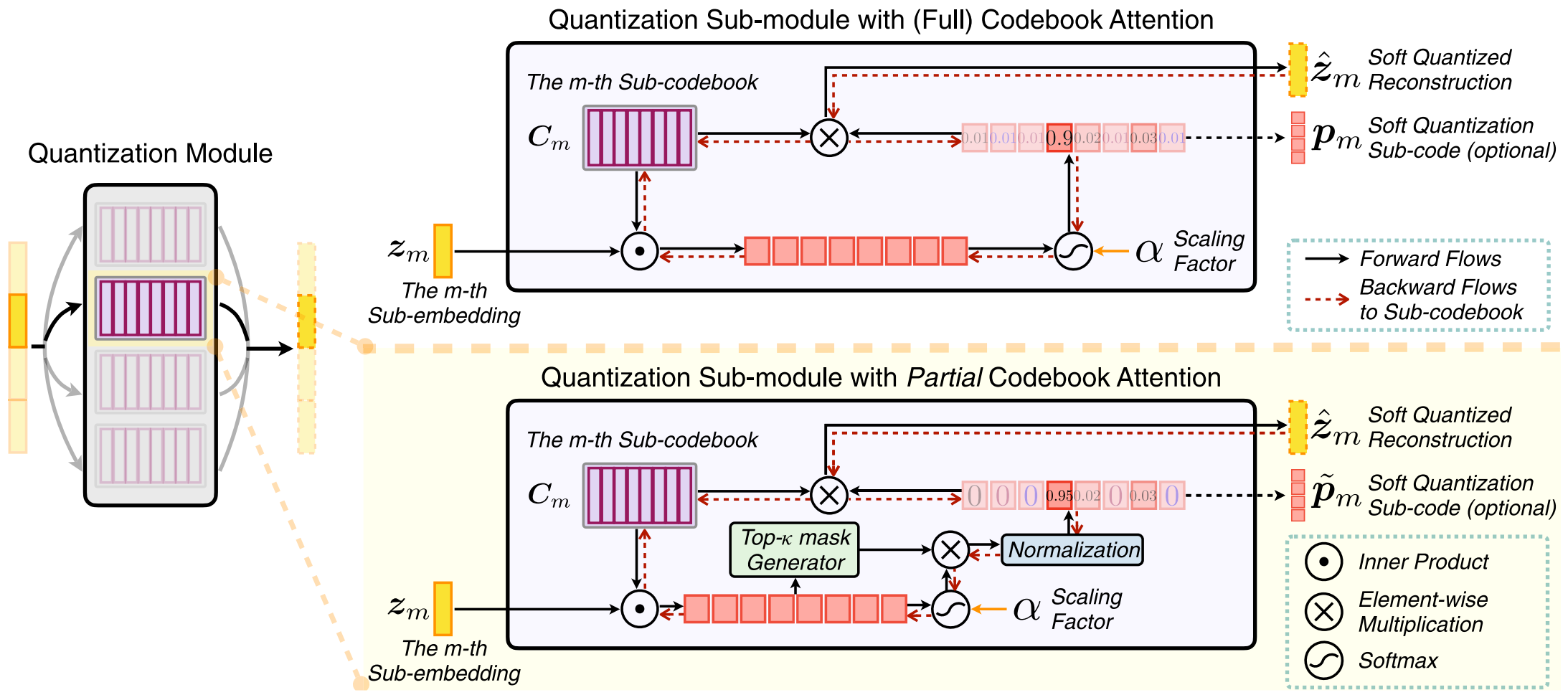

**Fig. 3.** Top: Existing quantization module with full codebook attention. Bottom: The proposed learnable quantization module with the *partial codebook attention mechanism* in PHPQ. In the $m$th quantization sub-module, we generate a mask $\boldsymbol{\pi}_m$ for the sub-embedding $\boldsymbol{z}_m$ by ranking all the attention scores in $\boldsymbol{p}_m$. Then we use the mask to select the top-$\kappa$ entries in $\boldsymbol{p}_m$ and disable the rest of the entries. Finally, we get the soft assigned codeword for $\boldsymbol{z}_m$ by partial attentive pooling with the $\boldsymbol{C}_m$ and the masked attention scores $\tilde{\boldsymbol{p}}_m$.

To address the shortcomings, Generalized Spatial Pooling (GSP) [25] is proposed. Given an input image $\boldsymbol{I}$, the feature map produced by one convolutional layer is denoted as $\boldsymbol{F} \in \mathbb{R}^{H\times W\times C}$, where $\boldsymbol{F}^i \in \mathbb{R}^{H\times W}$ is the tensor of the $i$-th channel. Let $a^i_\rho$ be the pooled output of the $i$-th channel and $\boldsymbol{f} \in \mathbb{R}^C$ be the descriptor generated by GSP, then we have

$$a^i_\rho = \frac{1}{HW}\left(\sum_{h\in H}\sum_{w\in W} {\boldsymbol{F}^i_{hw}}^{\rho}\right)^{\frac{1}{\rho}}, \tag{1}$$

$$\boldsymbol{f} = \mathrm{GSP}_\rho(\boldsymbol{F}) = [a^1_\rho, a^2_\rho, \ldots, a^C_\rho], \tag{2}$$

where $\rho$ is the focus factor of GSP. Note that when $\rho = 1$, GSP is equivalent to GAP and acts as GMP when $\rho \to +\infty$. A larger $\rho$ leads to smaller high-response areas.

To better understand how GSP works, we analyze the gradient of pooled output $a^i_\rho$ *w.r.t.* $\boldsymbol{F}^i_{hw}$:

$$\begin{aligned}\frac{\partial a^i_\rho}{\partial \boldsymbol{F}^i_{hw}} &= \frac{1}{HW}\left(\sum_{h'\in H}\sum_{w'\in W} {\boldsymbol{F}^i_{h'w'}}^{\rho}\right)^{\frac{1}{\rho}-1}\cdot {\boldsymbol{F}^i_{hw}}^{\rho-1}\\ &\propto \frac{{\boldsymbol{F}^i_{hw}}^{\rho-1}}{\min_{h'\in H, w'\in W} {\boldsymbol{F}^i_{h'w'}}^{\rho}}.\end{aligned} \tag{3}$$

It can be observed that the gradient exhibits an exponential growth with an exponent of $\rho - 1$ as the region response value $\boldsymbol{F}^i_{hw}$ increases. Increasing $\rho$ results in smaller gradients for low-response regions compared to high-response ones, effectively restricting the gradients in low-response areas relative to high-response areas. Therefore, a larger $\rho$ can highlight high response areas that contain discriminative parts and in the meantime restrain the gradients to low response areas, *e.g.*, the background. Hence by selecting a proper $\rho$, GSP can capture various discriminative details simultaneously while filtering out noises.

**Hybrid Pooling**: CNNs naturally provide multi-scale features because of their inherent pyramidal structures. To take advantage of such features, we propose a Pyramid Hybrid Pooling (PHP) module. Specifically, we first extract the feature maps from the $i$th stage of a ResNet-18 as $\boldsymbol{X}_{s_i}$, $i \in \{2, 3, 4\}$, respectively. Then, we apply GSP with different focus factors $\{\rho_{s_i}\}^4_{i=2}$ to $\boldsymbol{X}_{s_i}$, $i \in \{2, 3, 4\}$, respectively:

$$\boldsymbol{f}_{s_i} = \mathrm{GSP}_{\rho_{s_i}}(\boldsymbol{X}_{s_i}),\ i \in \{2, 3, 4\}, \tag{4}$$

Empirically, the feature maps at lower stages contain information of visual details while the feature maps at higher stages contain abstract semantic information. We control the focus factors $\{\rho_{s_i}\}^4_{i=2}$ to satisfy $\rho_{s_i} > \rho_{s_{i+1}}$ such that the model can capture fine-grained discriminative details in shallow layers while preserving high-level semantic information in deeper layers.

Next, we fuse multi-level GSP descriptors by

$$\boldsymbol{h}_{s_2} = \mathrm{FC}(\boldsymbol{f}_{s_2}), \tag{5}$$

$$\boldsymbol{h}_{s_3} = \mathrm{FC}(\boldsymbol{h}_{s_2} + \boldsymbol{f}_{s_3}), \tag{6}$$

$$\boldsymbol{h}_{s_4} = \boldsymbol{h}_{s_3} + \boldsymbol{f}_{s_4}, \tag{7}$$

where FC denotes a fully connected layer without activation. Finally, we apply a transformation layer $g : \mathbb{R}^{D_{s_4}} \mapsto \mathbb{R}^D$ for the linear fused representation $\boldsymbol{h}_{s_4}$ and get the image embedding:

$$\boldsymbol{z} = g(\boldsymbol{h}_{s_4}) \in \mathbb{R}^D. \tag{8}$$

### 3.3. Quantization module

**Product Quantization and Soft Quantization**: Product quantization [4] has been widely applied in large-scale retrieval systems [26]. Given an image embedding $\boldsymbol{z} \in \mathbb{R}^D$, we first split it into $M$ equal-length sub-vectors, *i.e.*, $\boldsymbol{z} = [\boldsymbol{z}_1, \boldsymbol{z}_2, \ldots, \boldsymbol{z}_M]$, $\boldsymbol{z}_m \in \mathbb{R}^d$, $d = D/M$. Then we quantize these sub-vectors by vector quantization independently, *i.e.*, assigning the closest codeword to the input sub-vector. This process for the $m$th sub-vector is formulated as:

$$q_m(\boldsymbol{z}_m) = \boldsymbol{c}_m^{k^*}, \text{ s.t. } k^* = \arg\min_k \left\| \boldsymbol{z}_m - \boldsymbol{c}_m^k \right\|_2, \tag{9}$$

where $q_m(\cdot)$ represents the sub-vector quantizer for the $m$th sub-vector and $\boldsymbol{c}_m^k \in \mathbb{R}^d$ denotes the $k$th sub-codewords in the $m$th sub-codebook, $\boldsymbol{C}_m \in \mathbb{R}^{K\times d}$. The whole codebook is derived from the cartesian product of the $M$ sub-codebooks, *i.e.*, $\boldsymbol{C} = \boldsymbol{C}_1 \times \cdots \times \boldsymbol{C}_M$. In this way, the embedding space can be divided into $K^M$ disjoint cells and the embeddings in each cell can be represented by the cell centroid. Hence, the pairwise distance (similarity) can be approximately measured by the inter-centroid distance (similarity).

Note that such a clustering-based step cannot be trained by back-propagation, so it is hard to be integrated into the deep learning framework. To tackle this issue, we adopt the soft quantization scheme proposed in [7]. Firstly, we apply $\ell_2$ normalization to all codewords and sub-vectors:

$$\boldsymbol{c}_m^k \leftarrow \boldsymbol{c}_m^k / \|\boldsymbol{c}_m^k\|_2, \quad \boldsymbol{z}_m \leftarrow \boldsymbol{z}_m / \|\boldsymbol{z}_m\|_2. \tag{10}$$

Then we calculate the attention scores on each sub-codebook:

$$\boldsymbol{p}_m = [p_m^1, p_m^2, \ldots, p_m^K] \in \mathbb{R}^K, \tag{11}$$

$$p_m^k = \frac{e^{2\alpha\langle \boldsymbol{z}_m, \boldsymbol{c}_m^k\rangle}}{\sum_{k'=1}^K e^{2\alpha\langle \boldsymbol{z}_m, \boldsymbol{c}_m^{k'}\rangle}}, m \in \{1, 2, \ldots, M\}, \tag{12}$$

where $\langle\cdot,\cdot\rangle$ denotes the inner-product, $\alpha$ is a scaling factor.

Next, we aggregate the sub-codewords in each sub-codebook according to the codebook attention scores:

$$\hat{\boldsymbol{z}}_m = \boldsymbol{p}_m \boldsymbol{C}_m = \sum_{k=1}^K p_m^k \boldsymbol{c}_m^k. \tag{13}$$

Finally, the soft quantized reconstruction embedding $\hat{\boldsymbol{z}}$ *w.r.t.* the original image embedding $\boldsymbol{z}$ is concatenated by

$$\hat{\boldsymbol{z}} = [\hat{\boldsymbol{z}}_1, \hat{\boldsymbol{z}}_2, \ldots, \hat{\boldsymbol{z}}_M]. \tag{14}$$

Since the softmax operator is differentiable, the soft quantization is learnable by the back-propagation.

**Partial Codebook Attention for Soft Quantization**: Although the soft quantization mechanism enables learnable quantization, it leads to a gap between soft quantization in training and hard quantization in inference. To mitigate this gap, existing method [7] adopts a large $\alpha$ in the codebook attention and makes the attention scores approximate to the one-hot distribution. However, simply enlarging $\alpha$ will increase the difficulty of updating codeword selection, because the optimizer adheres to the codeword that produces the peak value, potentially causing the quantization model to become trapped in a sub-optimal solution. On the other hand, a smaller $\alpha$ helps to optimize the model, but it fails to exclude irrelevant codewords, *i.e.*, the codewords that are associated with the least attention weights. Since the attention weights for irrelevant codewords keep non-zero, these codewords will contribute to the soft quantized representations (Eq. (13)) and also be updated in backpropagation. Hence they will always be perturbed by the "noisy" gradients, which degrade quantization and harm fine-grained retrieval.

To improve such issues, we propose a *Partial Codebook Attention Mechanism* to improve learnable quantization. As shown in Fig. 3, the partial codebook attention mechanism meticulously selects top-$\kappa$ relevant codewords for soft reconstruction according to the attention score (*i.e.*, distance) of each sub-embedding, thereby facilitating training efficiency. First, for the $m$-th sub-codebook, we sort the sub-codewords' attention scores obtained from Eq. (12) in descending order, denoted as $\mathcal{P}_m \in \mathbb{R}^K$, where $\mathcal{P}_m^{k,o_k}$ represents the $k$-th largest attention score, corresponding to the $o_k$-th sub-codeword. Next, we gather the indexes of the codewords associated with the top-$\kappa$ attention scores, where $\kappa$ is a threshold parameter, and create an attention mask $\boldsymbol{\pi}_m = [\pi_m^1, \ldots, \pi_m^K]$ for the $m$-th sub-codebook:

$$\pi_m^{o_k} = \begin{cases} 1, & k \le \kappa \\ 0, & \kappa < k \le K \end{cases} \tag{15}$$

Next, we multiply $\pi_m$ with the original weight vector $\boldsymbol{p}_m$ to disable the irrelevant scores:

$$\boldsymbol{p}_m \leftarrow \boldsymbol{p}_m \otimes \boldsymbol{\pi}_m, \tag{16}$$

where $\otimes$ denotes element-wise product operation.

After that, we normalize $\boldsymbol{p}_m$ and get refined scores $\tilde{\boldsymbol{p}}_m$ by

$$\tilde{p}_m^k = p_m^k / (\sum_{k'=1}^K p_m^k). \tag{17}$$

Finally, we rewrite Eq. (13) as

$$\hat{\boldsymbol{z}}_m = \tilde{\boldsymbol{p}}_m \boldsymbol{C}_m = \sum_{k=1}^K \tilde{p}_m^k \boldsymbol{c}_m^k. \tag{18}$$

By Eq. (15)–(18), only codewords that share top-$\kappa$ similarity with original data will be considered, while others are regarded as irrelevant ones and will be ignored. From another point of view, partial codebook attention acts as a filter and removes redundant noise (*e.g.*, the wrong class information) contained in irrelevant codewords. Therefore, it helps to retain the fine-grained discriminative features during soft reconstruction.

### 3.4. Loss function

We follow [10] and apply a Soft Reconstruction-based Cross-Entropy Loss (SR-CEL) to train the model, namely,

$$\mathcal{L}_{SR-CEL} = \frac{1}{N}\sum_{n=1}^N \left( -\log \frac{e^{o_{ny_n}/\tau}}{\sum_{c'=1}^{N_c} e^{o_{nc'}/\tau}} \right), \tag{19}$$

where $N$, $N_c$ represent the mini-batch size and the number of categories respectively, $o_{nc}$ denotes the logit of the classifier that the $n$-th image belongs to the $c$-th category, $y_n$ is the ground-truth label for the $n$th image, and $\tau$ denotes the temperature hyper-parameter for the classifier.

Besides, to learn compact intra-class representations and discriminative inter-class representations, we adopt the Contrastive Loss (CL) [27], namely,

$$d^+ = \frac{1}{|\mathcal{B}_c|^2} \sum_{\hat{\boldsymbol{z}}\in\mathcal{B}_c} \sum_{\hat{\boldsymbol{z}}^+\in\mathcal{B}_c\setminus\{\hat{\boldsymbol{z}}\}} \|\hat{\boldsymbol{z}} - \hat{\boldsymbol{z}}^+\|_2, \tag{20}$$

$$d^- = \frac{1}{|\mathcal{B}_c|(|\mathcal{B}| - |\mathcal{B}_c|)} \sum_{\hat{\boldsymbol{z}}\in\mathcal{B}_c} \sum_{\hat{\boldsymbol{z}}^-\in\mathcal{B}\setminus\mathcal{B}_c} \|\hat{\boldsymbol{z}} - \hat{\boldsymbol{z}}^-\|_2, \tag{21}$$

$$\mathcal{L}_{CL} = \max(d^+ - m^+, 0) + \max(m^- - d^-, 0), \tag{22}$$

where $\mathcal{B}$ denotes the set of soft quantized reconstructions for the images in the current training mini-batch, $\mathcal{B}_c$ denotes a specific subset of $\mathcal{B}$ that is associated with the $c$-th category. $m^+$ and $m^-$ are the pre-defined positive and negative margins respectively.

Therefore, the overall loss function of PHPQ is

$$\mathcal{L}_{PHPQ} = \mathcal{L}_{SR-CEL} + \gamma \mathcal{L}_{CL}, \tag{23}$$

where $\gamma$ is a hyper-parameter to balance two loss terms.

### 3.5. Encoding and retrieval

In the inference, the database images are encoded into a series of codeword indexes through hard quantization. Given an image $\boldsymbol{I}$, we first obtain the image embedding $\boldsymbol{z} = [\boldsymbol{z}_1, \dots, \boldsymbol{z}_M]$. The semantic index $\boldsymbol{i} = [i_1, \dots, i_M]$ is computed by searching the nearest codewords:

$$i_m = \arg\max_k \langle \boldsymbol{z}_m, \boldsymbol{c}_m^k \rangle, \quad k \in \{1, 2, \dots, K\}. \tag{24}$$

In Eq. (24), each $\boldsymbol{i}$ only takes $M \log_2 K$ bits to be stored, which implies the high storage efficiency of quantization.

As for retrieval, we use Asymmetric Quantizer Distance (AQD) [4]. Given a query $\boldsymbol{I}^q$ and a database item $\boldsymbol{I}^d$, AQD computes their similarity by:

$$\mathrm{AQD}(\boldsymbol{I}^q, \boldsymbol{I}^d) = \sum_{m=1}^{M} \langle \boldsymbol{v}_m^q, \boldsymbol{c}_m^{i_m^d} \rangle. \tag{25}$$

To accelerate similarity computation, we can pre-compute a query-to-codeword lookup table. Concretely, we define $\Xi_m^q \in \mathbb{R}^{M \times K}$ as the lookup table for the $m$-th sub-codebook, in which $\Xi_m^{q,i_m^d} = \langle \boldsymbol{v}_m^q, \boldsymbol{c}_m^{i_m^d} \rangle$. After obtaining $\Xi_m^q$, we can compute the similarity by summing up $M$ look-up entries:

$$\mathrm{AQD}(\boldsymbol{I}^q, \boldsymbol{I}^d) = \sum_{m=1}^{M} \Xi_m^{q,i_m^d}. \tag{26}$$

## 4. Experiments

### 4.1. Experimental setup

**Datasets**: We conduct experiments on two fine-grained benchmarks: **(i) CUB-200-2011** [8] is a fine-grained bird dataset comprising 11,788 images of 200 categories. The training set contains 5994 images, while the test set contains 5794 images. The database is identical to the training set. **(ii) Stanford Dogs** [9] is a fine-grained dog dataset comprising 20,580 images in 120 categories. There are about 150 images per category, of which 100 are picked to form the training set, and the rest are used as the test set. We adopt the official split in [10,21] and take the respective training set as the database.

**Evaluation Metrics**: Three metrics in previous works [10,21,28] are adopted to measure the performance of the proposed PHPQ, *i.e.*, the Mean Average Precision (**MAP**), the Precision-Recall curves (**P-R**), and the Precision curves *w.r.t.* different numbers of top returned items (**P@N**). The MAP is defined as:

$$\mathrm{MAP} = \frac{1}{Q} \sum_{i=1}^{Q} \frac{1}{N_{\mathrm{rel}(i)}} \sum_{n=1}^{N_D} \frac{N_{\mathrm{rel}(i)}^n}{n} \cdot \mathbb{I}(k), \tag{27}$$

where $Q$ denotes the total number of queries, $N_D$ denotes the database size, $N_{\mathrm{rel}(i)}$ denotes the total number of relevant items *w.r.t.* the $i$-th query, $N_{\mathrm{rel}(i)}^n$ represents the number of relevant images in top-$n$ returned results. $\mathbb{I}(k)$ is an indicator function, which is equal to 1 once the $k$-th returned image shares the same category with the query, otherwise $\mathbb{I}(k) = 0$.

**Compared Baselines**: **(i)** 12 coarse-grained hashing methods based on ResNet-18, LSH [29], SH [30], PQ [4], ITQ [31], OPQ [5], DPSH [32], DTH [33], DSH [6], HashNet [28], ADSH [34], PQN [7], and CSQ [17]. **(ii)** 6 fine-grained hashing methods based on ResNet-18, FPH [21], SRLH [10], CFH [11], DSaH [35], ExchNet [12], and sRLH [36]. **(iii)** 5 fine-grained hashing methods based on ResNet-50, ExchNet [12], $A^2$-Net [37], SEMICON [38], SPH [39], and DFMH [40].

**Implementation Details**: We adopt two pre-trained backbones, *i.e.*, ResNet-18/50 for a fair comparison. All ablation studies are conducted with ResNet-18 as the backbone unless specifically claimed. The maximum training epoch is set to be 70 with batch size 64. We adopt Adam [41] as the optimizer. The initial learning rate is set to be 0.0001. We employ a balanced sampling strategy that guarantees each batch consists of $m$ categories, with $n$ samples per category. Consequently, the positive pairs in Eq. (22) are derived from the $n-1$ samples within the same category, while the negative pairs originate from the remaining $(m-1) * n$ samples that belong to different categories. Other parameters are listed below: **(i)** The focus factor of GSP, $\rho_{s_2} = 3$, $\rho_{s_3} = 2$, and $\rho_{s_4} = 1$. **(ii)** The embedding dimension, $D = 1536$. **(iii)** The size of a sub-codebook, $K = 256$, while the number of codebook $M$ varies in $\{2, 4, 6, 8\}$ such that each code takes $M \log_2 K$ bits. **(iv)** The scaling factor for soft quantization, $\alpha = 16$. **(v)** The temperature parameter for $\mathcal{L}_{SR-CEL}$, $\tau = 0.50$ for CUB-200-2011, and $\tau = 0.25$ for Stanford Dogs. **(vi)** The weight parameters for $\mathcal{L}_{PHPQ}$, $\gamma = 1$.

### 4.2. Results and analysis

**Results:** We report the MAP results in Table 1, which demonstrates that PHPQ consistently outperforms or achieves competitive performance compared to its competitors. Specifically, compared to CFH, the best ResNet-18 baseline, PHPQ achieves an average MAP increase of **3.33%** and **1.73%** on CUB-200-2011 and Stanford Dogs, respectively. As for ResNet-50 baselines, despite slightly lagging behind DFMH on longer bits, PHPQ is more robust and performs well on extremely short bits, *e.g.*, 16 bits. As for quantization learning, the partial codebook attention mechanism can optimize the codebook more effectively. Therefore, it produces better binary codes for fine-grained image retrieval and performs better than SOTA quantization methods, *e.g.*, PQ, OPQ, PQN, and CSQ. Furthermore, compared to other baselines, the performance gap of PHPQ between the two datasets is smaller. It indicates that the proposed PHP module improves the generalization ability to capture fine-grained features among different types of objects.

Fig. 4 shows P-R curves and P@N curves on CUB-200-2011 and Stanford Dogs under 16 bits. PHPQ consistently reaches higher precision under the same recall rate and finds more relevant images given a fixed returned list length, which is favorable because the users may pay more attention to the top-ranked results in a retrieval system.

**Quantization Formulation:** We first validate the rationale behind the soft product quantization scheme and design a vector quantization counterpart without dividing the entire vector into $M$ segments, by setting $M = 1$, namely PHP-VQ in Table 3. The performance of PHP-VQ is significantly lower than PHPQ due to the insufficient number of codewords for discrimination, highlighting the importance of dividing the vector into $M$ parts.

**Component Analysis**: To explore the contribution of each component, we design 7 variants of PHPQ: **(i)** $\mathrm{PHPQ}_{\text{last fc}}$ directly adopts the outputs of the last fully connected layer of ResNet-18 as image features. **(ii)** $\mathrm{PHPQ}_{\mathrm{GAP}}$ adopts global average pooling in the PHP module, *i.e.*, setting $\rho_{s_1} = \rho_{s_2} = \rho_{s_3} = 1$ in Eq. (4). **(iii)** $\mathrm{PHPQ}_{\mathrm{GMP}}$ adopts global maximum pooling in the PHP module, *i.e.*, setting $\rho_{s_1} = \rho_{s_2} = \rho_{s_3} = +\infty$ in Eq. (4). **(iv)** $\mathrm{PHPQ}_{\rho_{s_1} < \rho_{s_2} < \rho_{s_3}}$ adopts $\rho_{s_1} < \rho_{s_2} < \rho_{s_3}$ in the PHP module. We set $\rho_{s_1} = 1, \rho_{s_2} = 2, \rho_{s_3} = 3$ in Eq. (4). **(v)** $\mathrm{PHPQ}_{\text{full attn}}$ removes the partial codebook attention mechanism in the quantization module by setting $\kappa = 256$ in Eq. (15). **(vi)** $\mathrm{PHPQ}_{\text{w/o } \mathcal{L}_{CL}}$ removes $\mathcal{L}_{CL}$ in Eq. (23) and uses $\mathcal{L}_{SR-CEL}$ as the loss function. **(vii)** $\mathrm{PHPQ}_{4 \text{ layers}}$ that incorporates features from stage 1, and we set $\rho_{s_1} = 4, \rho_{s_2} = 3, \rho_{s_3} = 2, \rho_{s_4} = 1$. According to the results in Table 2, we can learn the following conclusions.

*The PHP module helps to capture and preserve multi-level visual features.* PHPQ outperforms $\mathrm{PHPQ}_{\text{last fc}}$ by 1.90% and 1.18% on CUB-200-2011 and Stanford Dogs. It implies that the last FC layer of CNN is ineffective to capture fine-grained discriminative features, thus is not enough to be applied in fine-grained scenarios. Besides, PHPQ also outperforms $PHPQ_{GMP}$ and $PHPQ_{GAP}$. The reason can be explained by two-folds: (**i**) GMP has a relatively small activated area. It pays most attention to the highest response element, while missing other discriminative clues for fine-grained retrieval. (**ii**) GAP treats each element in a feature map equally. It fails to filter out noisy information from the background, thus producing less discriminative descriptors for fine-grained quantization. By contrast, the proposed PHP module shows superior performance to capture fine-grained semantic information, thus contributing to better quantized representations. Moreover, PHPQ outperforms $PHPQ_{\rho_{s_1}<\rho_{s_2}<\rho_{s_3}}$. It indicates that a descending order of focus factors helps to make the best of PHP module because a larger $\rho$ helps to capture visual details in lower stages while a small $\rho$ can better preserve high-level semantic information in higher stages. Finally, $PHPQ_{4\ layers}$ outperforms PHPQ, indicating that including more low-level features is beneficial for fine-grained representation learning. However, since the improvement is relatively minor, we continue to use features from three layers by default to reduce computational overhead.

**Table 1**
The MAP under different bits on CUB-200–2011 and Stanford Dogs. The underlined numbers denote absolute state-of-the-art.

| Method | CUB-200–2011 | | | | Stanford Dogs | | | |
|---|---|---|---|---|---|---|---|---|
| | 16 bits | 32 bits | 48 bits | 64 bits | 16 bits | 32 bits | 48 bits | 64 bits |
| *Coarse-grained* Hashing Methods (Backbone: ResNet-18) | | | | | | | | |
| LSH [29] | 8.67 | 9.14 | 18.57 | 19.31 | 8.01 | 19.54 | 30.99 | 41.62 |
| SH [30] | 26.39 | 43.90 | 51.39 | 55.12 | 25.26 | 43.69 | 49.69 | 55.66 |
| PQ [4] | 35.01 | 41.95 | 52.16 | 54.94 | 43.64 | 44.42 | 47.32 | 48.45 |
| ITQ [31] | 25.19 | 45.96 | 53.76 | 58.97 | 31.70 | 46.59 | 54.37 | 57.40 |
| OPQ [5] | 34.89 | 43.68 | 52.90 | 55.09 | 39.97 | 52.65 | 51.75 | 52.07 |
| DPSH [32] | 34.97 | 43.01 | 49.08 | 52.25 | 42.70 | 55.28 | 60.80 | 62.31 |
| DTH [33] | 46.41 | 54.54 | 57.71 | 58.81 | 54.35 | 62.58 | 63.62 | 65.73 |
| DSH [6] | 31.56 | 49.30 | 54.08 | 59.67 | 47.28 | 55.87 | 61.28 | 63.19 |
| HashNet [28] | 37.91 | 46.28 | 48.53 | 51.23 | 47.45 | 55.21 | 55.75 | 59.34 |
| ADSH [34] | 42.93 | 57.65 | 67.19 | 72.28 | 47.91 | 61.05 | 66.38 | 68.60 |
| PQN [7] | 59.07 | 65.89 | 70.39 | 72.91 | 65.94 | 67.04 | 70.76 | 70.95 |
| CSQ [17] | 54.41 | 63.93 | 67.61 | 69.68 | 59.45 | 67.96 | 70.66 | 71.92 |
| *Fine-grained* Hashing Methods (Backbone: ResNet-18) | | | | | | | | |
| FPH [21] | 51.69 | 58.32 | 61.24 | 62.33 | 63.40 | 69.09 | 70.60 | 71.30 |
| SRLH [10] | 69.08 | 69.22 | 70.87 | 70.10 | 66.68 | 74.02 | 75.16 | 75.38 |
| CFH [11] | 72.08 | 73.41 | 73.21 | 72.71 | 79.33 | 79.96 | 79.47 | 79.34 |
| DSaH [35] | 50.54 | 54.74 | 66.01 | 67.02 | 42.29 | 63.68 | 66.80 | 67.60 |
| ExchNet [12] | 51.04 | 65.02 | 68.56 | 70.03 | 61.92 | 71.48 | 72.03 | 72.16 |
| sRLH [36] | 62.68 | 69.37 | 71.27 | 71.60 | 65.27 | 72.32 | 74.36 | 75.33 |
| Ours | 76.03 | 76.24 | 76.25 | 76.21 | 81.01 | 81.27 | 81.44 | 81.29 |
| *Fine-grained* Hashing Methods (Backbone: ResNet-50) | | | | | | | | |
| ExchNet [12] | – | 67.74 | 71.05 | – | – | – | – | – |
| $A^2$-Net [37] | – | 71.61 | 77.33 | – | – | – | – | – |
| SEMICON [38] | – | 72.61 | 79.67 | – | – | – | – | – |
| SPH [39] | 79.19 | 84.26 | – | 85.76 | 76.64 | 82.02 | – | 82.20 |
| DFMH [40] | 81.79 | 85.66 | **86.98** | **87.51** | 83.39 | 85.35 | 86.02 | 86.45 |
| Ours | **85.70** | **86.08** | 86.97 | 86.36 | **87.19** | **87.28** | **87.42** | **87.41** |

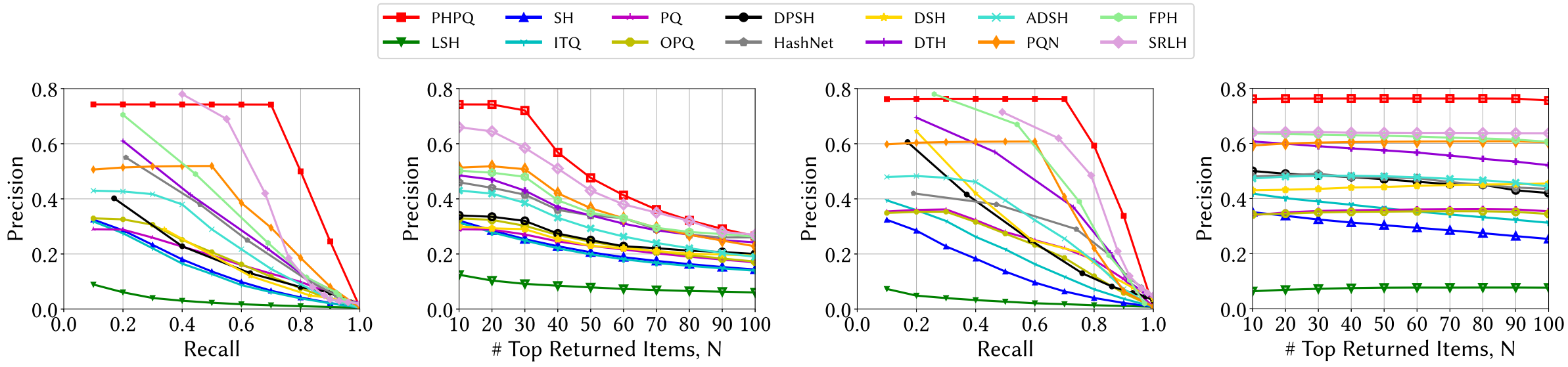

(a) CUB-200-2011 P-R Curve (b) CUB-200-2011 P@N Curve (c) Stanford Dogs P-R Curve (d) Stanford Dogs P@N Curve

**Fig. 4.** 16 bits Precision-Recall (P-R) and Precision@top-$N$ (P@N) curves on CUB-200-2011 and Stanford Dogs.

**Table 2**
The 16 bits MAP for different PHPQ variants on CUB-200–2011 and Stanford Dogs. "n/a" denotes not applicable.

| Variant | Features | Aggregation | $\{\rho_{s_i}\}_{i=1}^{3(4)}$ | $\mathcal{L}_{w/oCL}$ | $\kappa$ | CUB-200–2011 | Stanford Dogs |
|---|---|---|---|---|---|---|---|
| $PHPQ_{last\ fc}$ | ResNet | n/a | n/a | ✓ | 5 | 74.13 | 79.83 |
| $PHPQ_{GAP}$ | Pyramid | GAP | 1 | ✓ | 5 | 75.35 | 80.81 |
| $PHPQ_{GMP}$ | Pyramid | GMP | $+\infty$ | ✓ | 5 | 73.70 | 79.57 |
| $PHPQ_{\rho_{s_1}<\rho_{s_2}<\rho_{s_3}}$ | Pyramid | PHP | 1,2,3 | ✓ | 5 | 75.15 | 80.11 |
| $PHPQ_{full\ attn}$ | Pyramid | PHP | 3,2,1 | ✓ | 256 | 75.03 | 80.38 |
| $PHPQ_{w/o\mathcal{L}_{CL}}$ | Pyramid | PHP | 3,2,1 | × | 5 | 73.30 | 78.07 |
| PHPQ | Pyramid | PHP | 3,2,1 | ✓ | 5 | 76.03 | 81.01 |
| $PHPQ_{4\ layers}$ | Pyramid | PHP | 4,3,2,1 | ✓ | 5 | **76.13** | **81.15** |

**Table 3**
The MAP *w.r.t.* vector quantization and 16 bits product quantization on CUB-200–2011 and Stanford Dogs.

| Variant | Backbone | M, K | Codewords | Bits | CUB-200–2011 | Stanford Dogs |
|---|---|---|---|---|---|---|
| PHP-VQ | ResNet-18 | 1, 256 | 256 | 8 | 71.51 | 77.38 |
| PHPQ | | 2, 256 | $256^2$ | 16 | **76.03** | **81.01** |
| PHP-VQ | ResNet-50 | 1, 256 | 256 | 8 | 80.97 | 83.90 |
| PHPQ | | 2, 256 | $256^2$ | 16 | **84.70** | **87.19** |

**Table 4**
The MAP results *w.r.t.* different attention threshold $\kappa$s under different bits on CUB-200–2011 and Stanford Dogs.

| $\kappa$ | CUB-200–2011 | | | | Stanford Dogs | | | |
|---|---|---|---|---|---|---|---|---|
| | 16 bits | 32 bits | 48 bits | 64 bits | 16 bits | 32 bits | 48 bits | 64 bits |
| 1 | 67.41 | 73.47 | 74.51 | 75.03 | 50.56 | 74.76 | 77.91 | 79.55 |
| 5 | **76.03** | **76.24** | **76.25** | **76.21** | **81.01** | **81.27** | **81.44** | **81.29** |
| 25 | 75.87 | 75.98 | 76.15 | 76.12 | 80.79 | 80.97 | 81.23 | 81.05 |
| 125 | 75.14 | 75.82 | 76.03 | 76.03 | 80.53 | 80.54 | 80.87 | 80.59 |
| 256 | 75.03 | 75.74 | 75.99 | 75.98 | 80.38 | 80.42 | 80.67 | 80.45 |

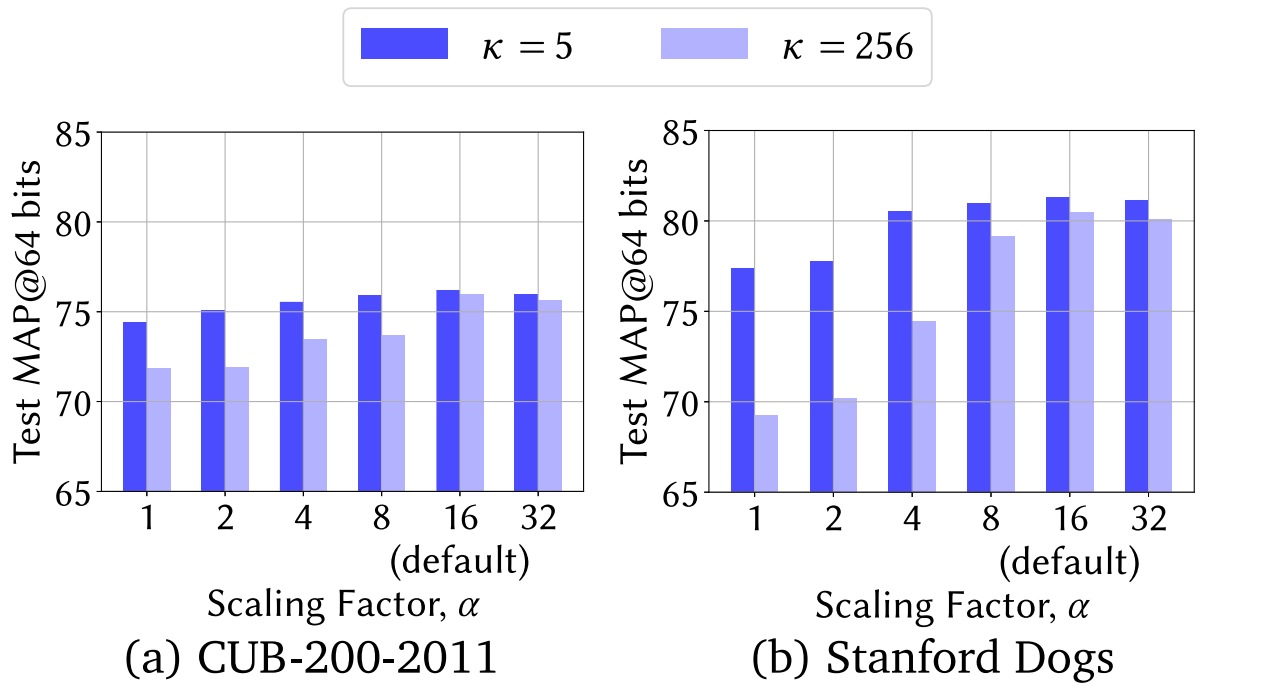

**Fig. 5.** The MAP results with partial codebook attention ($\kappa = 5$) and full codebook attention ($\kappa = K = 256$) *w.r.t.* different scaling factors in soft quantization. The models are evaluated under 64 bits.

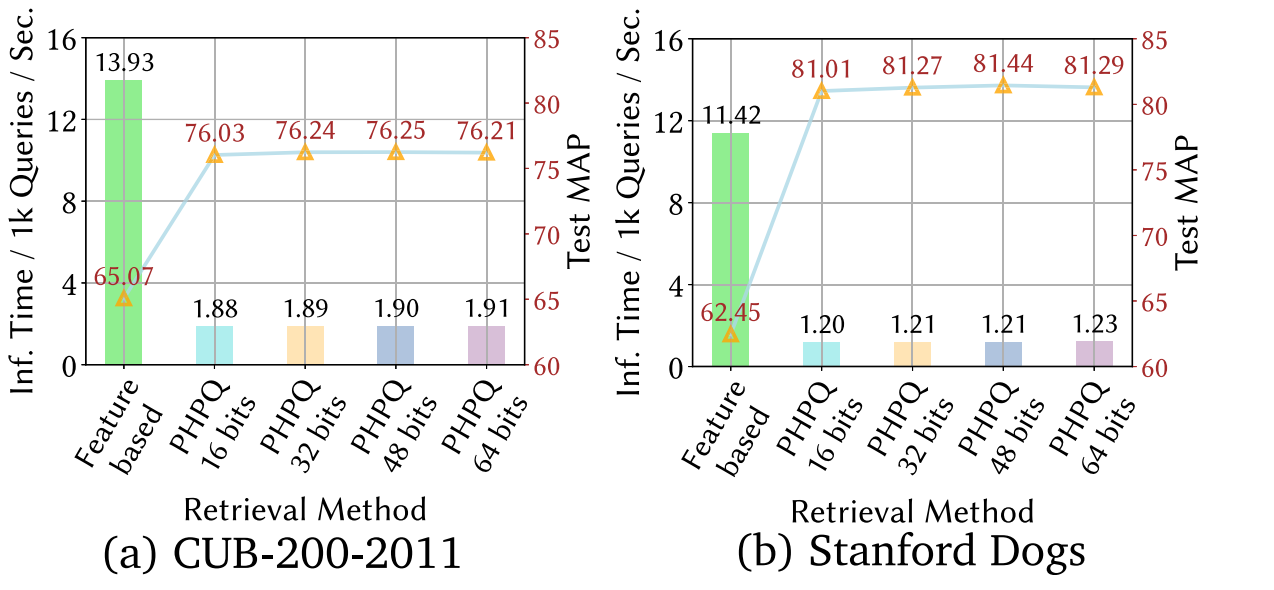

**Fig. 6.** The average inference time per $1 \times 10^3$ queries in seconds and the MAP results of different methods on CUB-200-2011 and Stanford Dogs.

**Table 5**
The parameters size, FLOPs and retrieval complexity. $N_q, N_{db}$ denote the query and database item numbers, respectively. $D$ denotes the embedding dimension, $L$ denotes the bit length.

| Method | Backbone | Params | GFLOPs | Retrieval Complexity |
|---|---|---|---|---|
| ResNet-18 | ResNet-18 | 12.0 M | 1.82 | $O(N_q * N_{db} * D)$ |
| sRLH$_{48\ bits}$ | | 12.3 M | 1.86 | $O(N_q * N_{db} * L)$ |
| PHPQ$_{48\ bits}$ | | 13.4 M | 1.96 | $O(N_q * K * D)$ |
| ResNet-50 | ResNet-50 | 26.7 M | 4.11 | $O(N_q * N_{db} * D)$ |
| SEMICON$_{48\ bits}$ | | 42.4 M | 7.22 | $O(N_q * N_{db} * L)$ |
| PHPQ$_{48\ bits}$ | | 32.0 M | 4.42 | $O(N_q * K * D)$ |

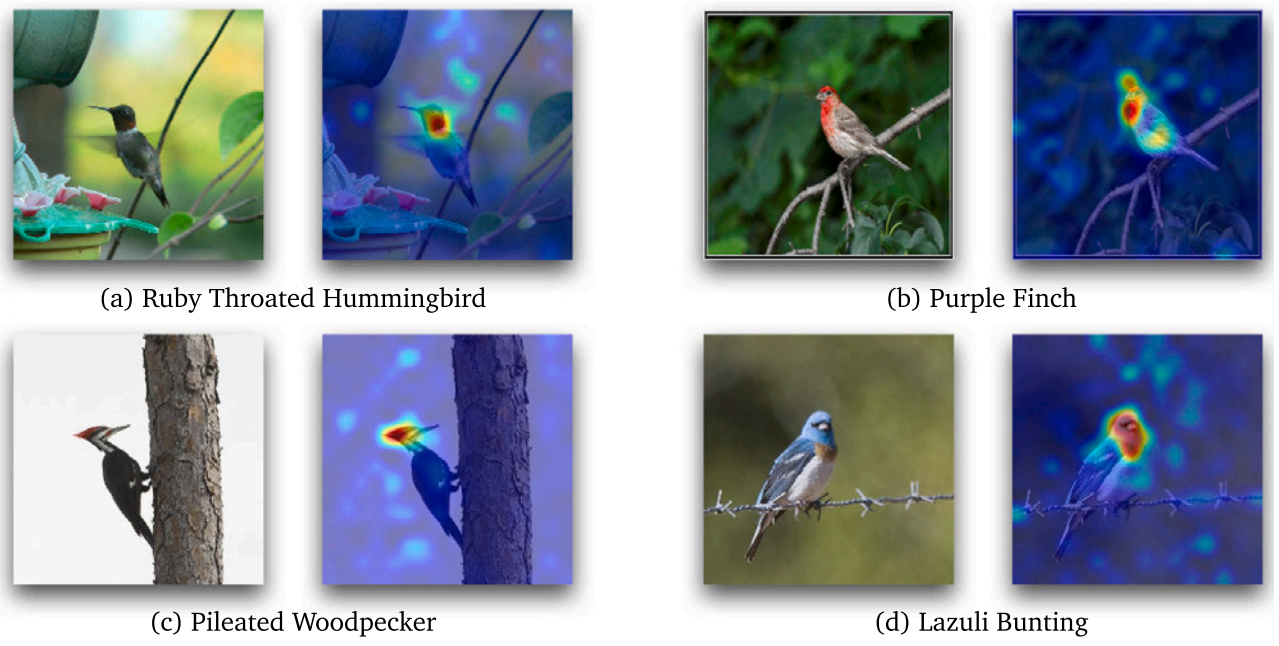

**Fig. 7.** The GradCAM [42] visualization of PHPQ on CUB-200-2011.

*Partial codebook attention improves the quantization learning.* PHPQ performs better than PHPQ $_{\text{full attn}}$ on all settings, especially at low-bit representations. Note that the scaling factor $\alpha$ is also an important factor in quantization training, we further investigate the relation between $\alpha$ and the partial codebook attention mechanism. As shown in Fig. 5, the partial codebook attention can consistently improve the quantization learning under different $\alpha$s, which indicates that partial codebook attention can be a universal way to improve trainable quantization. Besides, the large $\alpha$ also makes the model more difficult to be optimized and converge. By contrast, with the help of partial codebook attention, the model can achieve comparable performance with the best one even if $\alpha$ is relatively small. Hence the training efficiency is boosted without losing much precision.

We also explore the selection of the partial codebook attention threshold $\kappa$. As shown in Table 4, the model performs best under $\kappa = 5$. A tight threshold increases the risk of missing relevant codewords in the optimization. For example, the model performance under $\kappa = 1$ is not satisfactory. On the other side, a loose threshold is not favorable either, because it reduces the filtering effect of partial codebook attention. For instance, when $\kappa = K = 256$, the model achieves suboptimal performance compared with other settings, *i.e.*, $\kappa = 5, 25, 125$.

*Contrastive loss helps to learn discriminative quantization representation.* PHPQ outperforms PHPQ $_{\text{w/o}\ \mathcal{L}_{CL}}$. It shows that $\mathcal{L}_{CL}$ plays an assistant role to improve the intra-class compactness and inter-class diversity, thus making the classification borders clearer and improving retrieval performance.

**Efficiency Analysis**: Fig. 6 shows the inference time of different methods *w.r.t.* $1 \times 10^3$ queries in seconds and their MAP results on two datasets. The average speed of PHPQ is **7.35** and **9.41** times faster than feature-based retrieval on CUB-200-2011 and Stanford Dogs respectively, which demonstrates the efficiency of our approach. Interestingly, we find a similar phenomenon as [7] that the quantization-based methods even produce better retrieval results than the original feature-based method. The reason is that quantization acts as an additional regularization on the outputs, thus suppresses overfitting [7].

Additionally, we compare the parameter size, FLOPs, and retrieval complexity in Table 5. In comparison to the ResNet baselines, our lightweight PHP modules and soft quantization introduce only a few additional parameters, keeping the GFLOPs nearly unchanged. Furthermore, we outperform SEMICON [38] with a 25% reduction in parameters, demonstrating the efficiency of PHPQ. For retrieval complexity, we calculate the multiplication time. Given $N_q$ queries, we only need to pre-compute the similarity between queries and codewords. This complexity, which is not dependent on the number of items, is particularly advantageous for large-scale scenarios.

**Visualization**: To demonstrate the superiority of PHPQ intuitively, we have used GradCAM [42] to visualize the activation map of PHPQ on CUB-200-2011. As shown in Fig. 7, the discriminative parts, such as the lazuli head, are captured precisely. This highlights the effectiveness of our proposed lightweight discriminative feature mining module, *i.e.*, PHP.

## 5. Conclusion

In this paper, we propose a deep quantization model for fine-grained image retrieval, namely ***Pyramid Hybrid Pooling Quantization*** (**PHPQ**). Concretely, we propose a *Pyramid Hybrid Pooling* module that captures and preserves fine-grained semantic information through generalized spatial pooling over multi-level features. Besides, we propose a learnable quantization module with a *Partial Codebook Attention Mechanism*, which helps to optimize the most relevant codewords and improves performance. Comprehensive experiments demonstrate the superiority of PHPQ over state-of-the-art baselines. The future work is two-fold: (**i**) An input-aware or class-aware pooling scheme for the PHP module. (**ii**) A dynamic partial codebook attention mechanism that adapts the number of candidate codewords during the training.

## CRediT authorship contribution statement

**Ziyun Zeng:** Conceptualization, Data curation, Formal analysis, Investigation, Methodology, Project administration, Supervision, Validation, Visualization, Writing – original draft, Writing – review & editing. **Jinpeng Wang:** Conceptualization, Data curation, Formal analysis, Investigation, Methodology, Project administration, Supervision, Validation, Visualization, Writing – original draft, Writing – review & editing. **Bin Chen:** Conceptualization, Data curation, Formal analysis, Investigation, Methodology, Project administration, Supervision, Validation, Visualization, Writing – original draft, Writing – review & editing. **Tao Dai:** Conceptualization, Data curation, Formal analysis, Investigation, Methodology, Project administration, Supervision, Validation, Visualization, Writing – original draft, Writing – review & editing. **Shu-Tao Xia:** Conceptualization, Data curation, Formal analysis, Investigation, Methodology, Project administration, Supervision, Validation, Visualization, Writing – original draft, Writing – review & editing. **Zhi Wang:** Conceptualization, Data curation, Formal analysis, Investigation, Methodology, Project administration, Supervision, Validation, Visualization, Writing – original draft, Writing – review & editing.

## Declaration of competing interest

The authors declare the following financial interests/personal relationships which may be considered as potential competing interests: Bin Chen reports financial support was provided by Guangdong Basic and Applied Basic Research Foundation. Bin Chen reports a relationship with Guangdong Basic and Applied Basic Research Foundation that includes: funding grants.

## Data availability

The dataset is available on the internet.

## Acknowledgment

This work is supported in part by the National Natural Science Foundation of China under grant 62171248, 62301189, Guangdong Basic and Applied Basic Research Foundation under grant 2021A15 15110066, Guangdong Provincial Key Laboratory of Novel Security Intelligence Technologies (2022B1212010005), and Shenzhen Science and Technology Program under Grant JCYJ20220818101012025, JCY J20220818101014030, RCBS20221008093124061, GXWD202208111 72936001, RCYX20200714114523079, and the PCNL KEY project (PC L2023AS6-1).

## References

[1] J. Wang, T. Zhang, N. Sebe, H.T. Shen, et al., A survey on learning to hash, IEEE Trans. Pattern Anal. Mach. Intell. (2018) http://dx.doi.org/10.1109/TPAMI.2017.2699960.
[2] Y. Xie, R. Wei, J. Song, Y. Liu, Y. Wang, K. Zhou, Label-affinity self-adaptive central similarity hashing for image retrieval, IEEE Trans. Multim. (2023) http://dx.doi.org/10.1109/TMM.2023.3248170.
[3] D. Shi, L. Zhu, J. Li, Z. Zhang, X. Chang, Unsupervised adaptive feature selection with binary hashing, IEEE Trans. Image Process. (2023) http://dx.doi.org/10.1109/TIP.2023.3234497.
[4] H. Jegou, M. Douze, C. Schmid, Product quantization for nearest neighbor search, IEEE Trans. Pattern Anal. Mach. Intell. (2010) http://dx.doi.org/10.1109/TPAMI.2010.57.
[5] T. Ge, K. He, Q. Ke, J. Sun, Optimized product quantization for approximate nearest neighbor search, in: CVPR, 2013, http://dx.doi.org/10.1109/CVPR.2013.379.
[6] H. Liu, R. Wang, S. Shan, X. Chen, Deep supervised hashing for fast image retrieval, in: CVPR, 2016, http://dx.doi.org/10.1109/CVPR.2016.227.
[7] T. Yu, J. Meng, C. Fang, H. Jin, J. Yuan, Product quantization network for fast visual search, Int. J. Comput. Vis. (2020) http://dx.doi.org/10.1007/S11263-020-01326-X.
[8] C. Wah, S. Branson, P. Welinder, P. Perona, S. Belongie, The Caltech-ucsd Birds-200–2011 Dataset, California Institute of Technology, 2011.
[9] A. Khosla, N. Jayadevaprakash, B. Yao, F.-F. Li, Novel dataset for fine-grained image categorization: Stanford dogs, in: CVPRW, 2011.
[10] H. Zeng, H. Lai, J. Yin, Simultaneous region localization and hash coding for fine-grained image retrieval, 2019, arXiv:1911.08028.
[11] L. Ma, X. Li, Y. Shi, J. Wu, Y. Zhang, Correlation filtering-based hashing for fine-grained image retrieval, IEEE Signal Process. Lett. (2020) http://dx.doi.org/10.1109/LSP.2020.3039755.
[12] Q. Cui, Q.-Y. Jiang, X.-S. Wei, W.-J. Li, O. Yoshie, ExchNet: A unified hashing network for large-scale fine-grained image retrieval, in: ECCV, 2020, http://dx.doi.org/10.1007/978-3-030-58580-8_12.
[13] J. Wang, Z. Zeng, B. Chen, T. Dai, S.-T. Xia, Contrastive quantization with code memory for unsupervised image retrieval, in: AAAI, 2022, http://dx.doi.org/10.1609/AAAI.V36I3.20147.
[14] Y. Chen, Z. Wang, Y. Peng, Z. Zhang, G. Yu, J. Sun, Cascaded pyramid network for multi-person pose estimation, in: CVPR, 2018, http://dx.doi.org/10.1109/CVPR.2018.00742.
[15] B. Liu, Y. Cao, M. Long, J. Wang, J. Wang, Deep triplet quantization, in: ACM Multimedia, 2018, http://dx.doi.org/10.1145/3240508.3240516.
[16] J. Wang, B. Chen, T. Dai, S.-T. Xia, Webly supervised deep attentive quantization, in: ICASSP, 2021, http://dx.doi.org/10.1109/ICASSP39728.2021.9414172.
[17] L. Yuan, T. Wang, X. Zhang, F.E. Tay, Z. Jie, W. Liu, J. Feng, Central similarity quantization for efficient image and video retrieval, in: CVPR, 2020, http://dx.doi.org/10.1109/CVPR42600.2020.00315.
[18] A. Krizhevsky, I. Sutskever, G.E. Hinton, Imagenet classification with deep convolutional neural networks, in: NIPS, 2012.
[19] K. He, X. Zhang, S. Ren, J. Sun, Deep residual learning for image recognition, in: CVPR, 2016, http://dx.doi.org/10.1109/CVPR.2016.90.
[20] A.v.d. Oord, Y. Li, O. Vinyals, Representation learning with contrastive predictive coding, 2018, arXiv:1807.03748.
[21] Y. Yang, L. Geng, H. Lai, Y. Pan, J. Yin, Feature pyramid hashing, in: ICMR, 2019, http://dx.doi.org/10.1145/3323873.3325015.
[22] Z. Wang, X. Shang, Spatial pooling strategies for perceptual image quality assessment, in: ICIP, 2006, http://dx.doi.org/10.1109/ICIP.2006.313136.
[23] G. Tolias, R. Sicre, H. Jégou, Particular object retrieval with integral max-pooling of cnn activations, in: ICLR, 2016.
[24] A. Babenko, V. Lempitsky, Aggregating local deep features for image retrieval, in: ICCV, 2015, http://dx.doi.org/10.1109/ICCV.2015.150.
[25] F. Radenović, G. Tolias, O. Chum, Fine-tuning CNN image retrieval with no human annotation, IEEE Trans. Pattern Anal. Mach. Intell. (2018) http://dx.doi.org/10.1109/TPAMI.2018.2846566.
[26] J. Johnson, M. Douze, H. Jégou, Billion-scale similarity search with GPUs, IEEE Trans. Big Data (2021) http://dx.doi.org/10.1109/TBDATA.2019.2921572.
[27] R. Hadsell, S. Chopra, Y. LeCun, Dimensionality reduction by learning an invariant mapping, in: CVPR, 2006, http://dx.doi.org/10.1109/CVPR.2006.100.
[28] Z. Cao, M. Long, J. Wang, P.S. Yu, Hashnet: Deep learning to hash by continuation, in: ICCV, 2017, http://dx.doi.org/10.1109/ICCV.2017.598.
[29] M. Datar, N. Immorlica, P. Indyk, V.S. Mirrokni, Locality-sensitive hashing scheme based on p-stable distributions, in: SCG, 2004, http://dx.doi.org/10.1145/997817.997857.
[30] Y. Weiss, A. Torralba, R. Fergus, et al., Spectral hashing, in: NIPS, 2008.
[31] Y. Gong, S. Lazebnik, A. Gordo, F. Perronnin, Iterative quantization: A procrustean approach to learning binary codes for large-scale image retrieval, IEEE Trans. Pattern Anal. Mach. Intell. (2012) http://dx.doi.org/10.1109/TPAMI.2012.193.

[32] W.-J. Li, S. Wang, W.-C. Kang, Feature learning based deep supervised hashing with pairwise labels, in: IJCAI, 2016.
[33] H. Lai, Y. Pan, Y. Liu, S. Yan, Simultaneous feature learning and hash coding with deep neural networks, in: CVPR, 2015, http://dx.doi.org/10.1109/CVPR.2015.7298947.
[34] Q.-Y. Jiang, W.-J. Li, Asymmetric deep supervised hashing, in: AAAI, 2018, http://dx.doi.org/10.1609/AAAI.V32I1.11814.
[35] S. Jin, H. Yao, X. Sun, S. Zhou, L. Zhang, X. Hua, Deep saliency hashing for fine-grained retrieval, IEEE Trans. Image Process. (2020) http://dx.doi.org/10.1109/TIP.2020.2971105.
[36] X. Xiang, Y. Zhang, L. Jin, Z. Li, J. Tang, Sub-region localized hashing for fine-grained image retrieval, IEEE Trans. Image Process. (2022) http://dx.doi.org/10.1109/TIP.2021.3131042.
[37] X.-S. Wei, Y. Shen, X. Sun, H.-J. Ye, J. Yang, $A^2$-Net: Learning attribute-aware hash codes for large-scale fine-grained image retrieval, in: NIPS, 2021.
[38] Y. Shen, X. Sun, X.-S. Wei, Q.-Y. Jiang, J. Yang, SEMICON: A learning-to-hash solution for large-scale fine-grained image retrieval, in: ECCV, 2022, http://dx.doi.org/10.1007/978-3-031-19781-9_31.
[39] H. Sun, Y. Fan, J. Shen, N. Liu, D. Liang, H. Zhou, A novel semantics-preserving hashing for fine-grained image retrieval, IEEE Access (2020) http://dx.doi.org/10.1109/ACCESS.2020.2970223.
[40] W. Lang, H. Sun, C. Xu, N. Liu, H. Zhou, Discriminative feature mining hashing for fine-grained image retrieval, J. Vis. Commun. Image Represent. (2022) http://dx.doi.org/10.1016/J.JVCIR.2022.103592.
[41] D.P. Kingma, J. Ba, Adam: A method for stochastic optimization, in: ICLR, 2015.
[42] J. Gildenblat, contributors, Pytorch library for CAM methods, 2021, GitHub, https://github.com/jacobgil/pytorch-grad-cam.